\documentclass{article}
\usepackage[utf8]{inputenc}
\usepackage[T1]{fontenc}
\usepackage[babel=true]{csquotes}
\usepackage{textcomp}
\usepackage{amsmath,amssymb}
\usepackage{cite}
\usepackage{graphicx}
\usepackage{array}
\usepackage{tabularx}
\usepackage{color}
\usepackage{algorithm,algorithmic}

\title{Random Forests for Industrial Device Functioning Diagnostics Using Wireless Sensor Networks}
\author{W. ElGhazel, C. Guyeux, A. Farhat, M. Hakem,\\ K. Medjaher, N. Zerhouni, and J.M. Bahi}
\date{}

\begin{document}
\maketitle

\abstract{In this paper, random forests are proposed for operating devices diagnostics in the presence of a variable number of features.
In various contexts, like large or difficult-to-access monitored areas, wired sensor networks providing features to achieve diagnostics are either very costly to use or totally impossible to spread out. Using a wireless sensor network can solve this problem, but this latter is more subjected to flaws. Furthermore, the networks' topology often changes, leading to a variability in quality of coverage in the targeted area. Diagnostics at the sink level must take into consideration that both the number and the quality of the provided features are not constant, and that some politics like scheduling or data aggregation may be developed across the network. The aim of this article is ($1$) to show that random forests are relevant in this context, due to their flexibility and robustness, and ($2$) to provide first examples of use of this method for diagnostics based on data provided by a wireless sensor network.}

\section{Introduction}

In machine learning, classification refers to identifying the class to which a new observation belongs, on the basis of a training set and quantifiable observations, known as properties.\\
In ensemble learning, the classifiers are combined to solve a particular computational intelligence problem. Many research papers encourage adapting this solution to improve the performance of a model, or reduce the likelihood of selecting a weak classifier.
For instance, Dietterich argued that averaging the classifiers' outputs guarantees a better performance than the worst classifier \cite{Dietterich00}. This claim was theoretically proven correct by Fumera and Roli \cite{Fumera05}. 
In addition to this, and under particular hypotheses, the fusion of multiple classifiers can improve the performance of the best individual classifier \cite{Tumer96}.

Two of the early examples of ensemble classifiers are Boosting and Bagging. 
In Boosting algorithm \cite{Schapire99}, the distribution of the training set changes adaptively based on the errors generated by the previous classifiers. In fact, at each step, a higher degree of importance is accorded to the misclassified instances. At the end of the training, a weight is accorded to each classifier, regarding its individual performance, indicating its importance in the voting process. As for Bagging \cite{Breiman96}, the distribution of the training set changes stochastically and equal votes are accorded to the classifiers. For both classifiers, the error rate decreases when the size of the committee increases.

In a comparison made by Tsymbal and Puuronen \cite{Tsymbal00}, it is shown that Bagging is more consistent but unable to take into account the heterogeneity of the instance space. In the highlight of this conclusion, the authors emphasize the importance of classifiers' integration. Combining various techniques can provide more accurate results as different classifiers will not behave in the same manner faced to some particularities in the training set. Nevertheless, if the classifiers give different results, a confusion may be induced \cite{Kanemoto13}. It is not easy to ensure reasonable results while combining the classifiers.
In this context, the use of random methods could be beneficial. Instead of combining different classifiers, a random method uses the same classifier over different distributions of the training set. A majority vote is then employed to identify the class.

In this article, the use of random forests (RF) is proposed for industrial functioning diagnostics, particularly in the context of devices being monitored using a wireless sensor network (WSN). A prerequisite in diagnostics is to consider that data provided by sensors are either flawless or simply noisy. However, deploying a wired sensor network on the monitored device is costly in some well-defined situations, specifically in large scale, moving, or hardly accessible areas to monitor. Such situations encompass nuclear power plants or any structure spread in deep water or in the desert. Wireless sensors can be considered in these cases, due to their low cost and easy deployment. 

WSNs monitoring is somehow unique in the sense that sensors too are subjected to failures or energy exhaustion, leading to a change in the network topology. Thus, monitoring quality is variable too and it depends on both time and location on the device. Various strategies can be deployed on the network to achieve fault tolerance or to extend the WSN's lifetime, like nodes scheduling or data aggregation. However, the diagnostic processes must be compatible with these strategies, and with a device coverage of a changing quality. The objective of this research work is to show that RF achieve a good compromise in that situation, being compatible with a number of sensors which may be variable over time, some of them being susceptible to errors. More precisely, we will explain why random methods are relevant to achieve accurate diagnostics of an industrial device being monitored using a WSN. The functioning of RF will then be recalled and applied in the monitoring context. Algorithms will be provided, and an illustration on a simulated WSN will finally be detailed. 

The remainder of this article is organized as follows. Section \ref{related work} summarizes the related work. In Section \ref{diagnostics}, we overview the research works in industrial diagnostics. We present the random forest algorithm in Section \ref{RF} and give simulation results in Section \ref{sim}. This research work ends with a conclusion section, where the contribution is summarized and intended future work is provided.

\section{Related work}
\label{related work}

Many research works have contributed in improving the classification's accuracy. For instance, tree ensembles use majority voting to identify the most popular class. They have the advantage of transforming weak classifiers into strong ones by combining their knowledge to reduce the error rate.

Usually, the growth of each tree is governed by random vectors sampled from the training set, and bagging is one of the early examples of this. In this method, each tree is grown by randomly selecting individuals from the training set without replacing them \cite{Breiman96}. The use of bagging can be motivated by three main reasons: ($1$) it enhances accuracy with the use of random features, ($2$) it gives ongoing estimates of the generalization error, strength, and correlation of combined trees, and ($3$) it is also good for unstable classifiers with large variance.

Meanwhile, Freund introduced the adaptive boosting algorithm Adaboost, which he defined as \cite{Freund96}:
\enquote{a deterministic algorithm that selects the weights on the training set for input to the next classifier based on the wrong classifications in the previous classifiers}.\\
The fact that the classifier focuses on correcting the errors at each new step remarkably improved the accuracy of classifications.

Shortly after, in \cite{Amit97} randomness was again used to grow the trees. The split was defined at each node by searching for the best random selection of features in the training set. Ho \cite{Ho98} introduced the random subspace, in which he randomly selects a subset of vectors of features to grow each tree. Diettrich introduced the random split selection where at each node, a split is randomly selected among $k$ best splits \cite{Dietterich00}. 

For these methods, and like bagging, a random vector sampled to grow a tree is completely independent from the previous vectors, but is generated with the same distribution.

Random split selection \cite{Dietterich00} and introducing random noise into the outputs \cite{Breiman99} both gave better results than bagging. Nevertheless, the algorithms implementing ways of re-weighting the training set, such as Adaboost \cite{Freund96}, outperform these two methods \cite{Breiman01}.

Therefore, Breiman combined the strengths of the methods detailed above into the random forest algorithm. In this method, individuals are randomly selected from the training set with replacement. At each node, a split is selected by reducing the dispersion generated by the previous step and consequently lowering the error rate. This algorithm is further detailed in Section \ref{RF}.

\section{Overview of diagnostics}
\label{diagnostics}

With their constantly growing complexity, current industrial systems witness costly downtime and failures. Therefore, an efficient health assessment technique is mandatory. In fact, in order to avoid expensive shutdowns, maintenance activities are scheduled to prevent interruptions in system operation. In early frameworks, maintenance takes place either after a failure occurs (corrective maintenance), or according to predefined time intervals (periodic maintenance). Nevertheless, this still generates extra costs due to \enquote{too soon} or \enquote{too late} maintenances. Accordingly, considering the actual health state of the operating devices is important in the decision making process. Maintenance here becomes condition-based, and is only performed after the system being diagnosed in a certain health state.

Diagnostics is an understanding of the relationship between what we observe in the present and what happened in the past, by relating the cause to the effect. After a fault takes place, and once detected, an anomaly is reported in the system behavior. The fault is then isolated by determining and locating the cause (or source) of the problem. Doing so, the component responsible for the failure is identified and the extent of the current failure is measured. This activity should meet several requirements in order to be efficient \cite{Dash00}. these requirements are enumerated in the following.

\begin{itemize}

\item \textbf{Early detection:} in order to improve industrial systems' reliability, fault detection needs to be quick and accurate. Nevertheless, diagnostic systems need to find a reasonable trade-off between quick response and fault tolerance. In other words, an efficient diagnostic system should differentiate between normal and erroneous performances in the presence of a fault.

\item \textbf{Isolability:} fault isolation is a very important step in the diagnostic process. It refers to the ability of a diagnostic system to determine the source of the fault and identify the responsible component. With the isolability attribute, the system should discriminate between different failures. When an anomaly is detected, a set of possible faults is generated. While the completeness aspect requires the actual faults to be a subset of the proposed set, resolution optimization necessitates that the set is as small as possible. A tradeoff then needs to be found while respecting the accuracy of diagnostics.

\item \textbf{Robustness and resources:} it is highly desirable that the diagnostic system would degrade gracefully rather than fail suddenly. For this finality, the system needs to be robust to noise and uncertainties. In addition to this, a trade-off between system performance and computational complexity is to be considered. For example, on-line diagnostics require low complexity and higher storage capacities.

\item \textbf{Faults identifiability:} a diagnostics system is of no interest if it cannot distinguish between normal and abnormal behaviors. It is also crucial that not only the cause of every fault is identified, but also that new observations of malfunctioning would not be misclassified as a known fault or as normal behavior. While it is very common that a present fault leads to the generation of other faults, combining the effects of these faults is not that easy to achieve due to a possible non-linearity. On the other hand, modeling the faults separately may exhaust the resources in case of large processes.

\item \textbf{Clarity:} when diagnostic models and human expertise are combined together, the decision making support is more reliable. Therefore, it is appreciated that the system explains how the fault was triggered and how it propagated, and keeps track on the cause/effect relationship. This can help the operator use their experience to evaluate the system and understand the decision making process. 

\item \textbf{Adaptability:} operating conditions, external inputs, and environmental conditions change all the time. Thus, to ensure relevant diagnostics at all levels,the system should adapt to changes and evolve in the presence of new information.

\end{itemize}

Existent diagnostic models have several limitations. Some of which are summarized in Table \ref{diagnosis}.

\begin{table}[!ht]
\centering
\begin{tabular}{|l|p{6.5cm}|}
\hline
\textbf{Diagnostic model} & \textbf{Drawbacks} \\
\hline
\hline
Markovian process & -Aging is not considered \\
 & -Different stages of degradation process cannot be accounted for\\
 & -Large volume of data is required for the training \\
 & -The assumptions are not always practical\\
\hline
Bayesian networks & -Prior transitions are not considered \\
& -Complete reliance on accurate thresholds\\
& -Many state transitions are needed for efficient results\\
& -Unable to predict unanticipated states\\
\hline
Neural networks & -Significant amount of data for the training \\
& -Retraining is necessary with every change of conditions\\
& -Pre-processing is needed to reduce inputs\\
\hline
Fuzzy systems & -Increasing complexity with every new entry\\
 & -Domain experts are required\\
 & -Results are as good as the developers' understanding\\
\hline
\end{tabular}
\caption{Limitations of diagnostic models}
\label{diagnosis}
\end{table}

The degradation process can be considered as a stochastic process. The evolution of the degradation is a random variable that describes the different levels of the system's health state, from good condition to complete deterioration. The deterioration process is multistate and can be divided into two main categories \cite{Moghaddass14}:

\begin{enumerate}
\item Continuous-state space: the device is considered failed when the predefined threshold is reached. 
\item Discrete-state space: the degradation process is divided into a finite number of discrete levels.
\end{enumerate}

As condition-based maintenance relies on reliable scheduling of maintenance activities, an understanding of the degradation process is required. For this finality, in this paper, we consider the discrete-state space deterioration process.

\section{Random forests}
\label{RF}

The RF algorithm is mainly the combination of Bagging \cite{Breiman96} and random subspace \cite{Ho98} algorithms, and was defined by Leo Breiman as \enquote{a combination of tree predictors such that each tree depends on the values of a random vector sampled independently and with the same distribution for all trees in the forest}\cite{Breiman01}. This method resulted from a number of improvements in tree classifiers' accuracy.

This classifier maximizes the variance by injecting randomness in variable selection, and minimizes the bias by growing the tree to a maximum depth (no pruning). The steps of constructing the forest are detailed in Algorithm \ref{algo}.

\begin{algorithm}[H]
\caption{Random forest algorithm}
\label{algo}
\begin{algorithmic}
\REQUIRE Labeled training set $S$, Number of trees $T$, Number of features $F$.
\ENSURE Learned random forest $RF$.
    \STATE {\textbf{initialize} RF as empty}
    \FOR{$i$ in $1..T$}
        \STATE{$S_i'$ $\leftarrow$ bootstrap ($S$)}
        \STATE {\textbf{initialize} the root of tree $i$}
        \REPEAT
            \IF{current node is terminal}
                \STATE{\textbf{affect} a class}
                \STATE{\textbf{go to} the next unvisited node if any}
            \ELSE 
                 \STATE{\textbf{select} the best feature $f^*$ among $F$}
                 \STATE{sub-tree $\leftarrow$ split($S_i',f^*$)}
                 \STATE{\textbf{add} (leftChild, rightChild) to tree $i$}
            \ENDIF
        \UNTIL{all nodes are visited}
        \STATE{\textbf{add} tree i to the forest}
    \ENDFOR
\end{algorithmic}
\end{algorithm}

In a RF, the root of a tree $i$ contains the instances from the training subset $S_i'$, sorted by their corresponding classes. A node is terminal if it contains instances of one single class, or if the number of instances representing each class is equal. In the alternative case, it needs to be further developed (no pruning). For this purpose, at each node, the feature that guarantees the best split is selected as follows.

\begin{enumerate}

\item The information acquired by choosing a feature can be computed through:

\begin{enumerate}

\item The entropy of Shannon, which measures the quantity of information

\begin{equation}
Entropy (p)= -\sum_{k=1}^c P (k/p) \times \log (P(k/p))
\end{equation}

where $p$ is the number of examples associated to a position in the tree, $c$ is the total number of classes, $k/p$ denotes the fraction of examples associated to a position in the tree and labelled class $k$, $P(k/p)$ is the proportion of elements labelled class $k$ at a position $p$.

\item The Gini index, which measures the dispersion in a population

\begin{equation}
Gini(x)= 1- \sum_{k=1}^c P(k/p)^2
\end{equation}

where $x$ is a random sample, $c$ is the number of classes, $k/p$ denotes the fraction of examples associated to a position in the tree and labelled class $k$, $P(k/p)$ is the proportion of elements labelled class $k$ at a position $p$.

\end{enumerate}

\item The best split is then chosen by computing the gain of information from growing the tree at given position, corresponding to each feature as follows:

\begin{equation}
Gain(p, t)= f(p) - \sum_{j=1}^n P_j \times f(p_j)
\end{equation}

where $p$ corresponds to the position in the tree, $t$ denotes the test at branch $n$, $P_j$ is the proportion of elements at position $p$ and that go to position $p_j$, $f(p)$ corresponds to either $Entropy(p)$ or $Gini(p)$.\\
The feature that provides the higher Gain is selected to split the node.

\end{enumerate}

The optimal training of a classification problem can be NP-hard. Tree ensembles have the advantage of running the algorithm from different starting points, and this can better approximate the near-optimal classifier.

In his paper, Leo Breiman discusses the accuracy of random Forests. In particular, he gave proof that the generalized error, although different from one application to another, always has an upper bound and so random forests converge \cite{Breiman01}.

The injected randomness can improve accuracy if it minimizes correlation while maintaining strength. The tree ensembles investigated by Breiman use either randomly selected inputs or a combination of inputs at each node to grow the tree. These methods have interesting characteristics as:

\begin{itemize}
\item[-] Their accuracy is at least as good as Adaboost
\item[-] They are relatively robust to outliers and noise
\item[-] They are faster than bagging or boosting
\item[-] They give internal estimates of error, strength, correlation, and variable importance
\item[-] They are simple and the trees can be grown in parallel
\end{itemize}

There are four different levels of diversity which were defined in \cite{Sharkey97}, level $1$ being the best and level $4$ the worst.

\begin{itemize}
\item \textbf{Level} $1$: no more than one classifier is wrong for each pattern.
\item \textbf{Level} $2$: the majority voting is always correct.
\item \textbf{Level} $3$: at least one classifier is correct for each pattern.
\item \textbf{Level} $4$: all classifiers are wrong for some pattern.
\end{itemize}

RF can guarantee that at least level two is reached. In fact, a trained tree is only selected to contribute in the voting if it does better than random, i.e. the error rate generated by the corresponding tree has to be less than $0.5$, or the tree will be dropped from the forest \cite{Breiman01}.

In \cite{Verikas11}, Verikas \textit{et al.} argue that the most popular classifiers (Support Vector Machine SVM, MultiLayer Perceptron MLP, and Relevance Vector Machine RVM) provide too little insight about the variable importance to the derived algorithm. They compared each of these methodologies to the random forest algorithm to find that in most cases RF outperform other techniques by a large margin.

\section{Experimental study}
\label{sim}

\subsection{Data collection} 

In this paper, we consider two sets of experiments. The sensor network is constituted by 110 nodes, sensing respectively the levels of temperature (50 sensors), pressure (50), and humidity (10) on the industrial device under consideration.

\subsubsection{Set of experiment $1$}

In this set of experiments, we consider that no level of correlation is introduced betweent the different features. Moreover, we suppose that at time $t$:

\begin{itemize}
\item Under normal conditions, temperature sensors follow a Gaussian law of parameter $(20\times (1+0.005t),1)$, while these parameters are mapped to $(35,1)$ in case of a malfunction of the industrial device. Finally, these sensors return the value 0 when they break down.
\item The Gaussian parameters are $(5 \times (1+0.01t),0.3)$ 
when both the industrial device and the pressure sensors are in normal conditions. The parameters are changed to $(15,1)$ in case of industrial failure, while the pressure sensors return 1 when they are themselves broken down.
\item Finally, the 10 humidity sensors produce data following a Gaussian law of parameter $(52.5\times (1+0.001 t),12.5)$ when they are sensing a well-functioning device. These parameters are set to $(70,10)$ in case of device failure, while malfunctioning humidity sensors produce the value 0.
\end{itemize}

\subsubsection{Set of experiment $2$}
\label{set2}

For this set, a linear correlation is injected between the studied features.

\begin{itemize}
\item Under normal conditions, temperature sensors follow a Gaussian law of parameter $(20\times (1+0.005t),1)$, while these parameters are mapped to $(35,1)$ in case of a malfunction of the industrial device. Finally, these sensors return the value 0 when they break down.
\item When both the industrial device and the pressure sensors are in normal conditions, the value of pressure is computed as $(x \div 2 + 10)$, where $x$ is the value of temperature. The parameters are changed to $(15,1)$ in case of industrial failure, while the pressure sensors return 1 when they are themselves broken down.
\item For a well-functioning device, the 10 humidity sensors produce data in the form of $(x \times 525 + 12)$. These parameters are set to $(70,10)$ in case of device failure, while malfunctioning humidity sensors produce the value 0.
\end{itemize}

For both data sets, the probability that a failure occurs at time $t$ follows a Bernoulli distribution of parameter $t \div 35000$.

Five levels of functioning are attributed to each category of sensors, depending on the abnormality of the sensed data. These levels are defined thanks to 4 thresholds, which are 22.9, 24.5, 26, and 28 degrees for the temperature (a temperature lower than 22.9°C is normal, while a sensed value larger than 28°C is highly related to a malfunctioning), 5.99, 6.4, 7.9, and 9 bars for the pressure parameter, and finally 68, 80, 92, and 95 percents for the humidity.

Data is generated as follows.

\begin{itemize}
\item For each time unit $t=1..100$ during the industrial device monitoring,

\begin{itemize}
\item For each category $c$ (temperature, pressure, humidity) of sensors:

\begin{itemize}
\item For each sensor $s$ belonging to category $c$:

\begin{itemize}
\item If $s$ has not yet detected a device failure:

\begin{enumerate}
\item $s$ picks a new data, according to the Gaussian law corresponding to a well-functioning device, which depends on both $t$ and $c$,
\item a random draw from the exponential law detailed previously is realized, to determine if a breakdown occurs on the location where $s$ is placed.
\end{enumerate}

\item Else $s$ picks a new datum according to the Bernoulli distribution of a category $c$ sensor observing a malfunctioning device.
\end{itemize}

\end{itemize}

\end{itemize}

\end{itemize}

The global failure level $F^t$ of a set of 110 sensed data produced by the wireless sensor network at a given time $t$ is defined as follows. For each sensed datum $d_i^t, i=1..110$, let $f_i^t \in \{1,..,5\}$ be the functioning level related to its category (pressure, temperature, or humidity). Then $F^t=\max{f_i^t\mid i=1..110}$.

\subsection{Random forest design}

\begin{figure}[ht]
\centering
\includegraphics[scale=0.25]{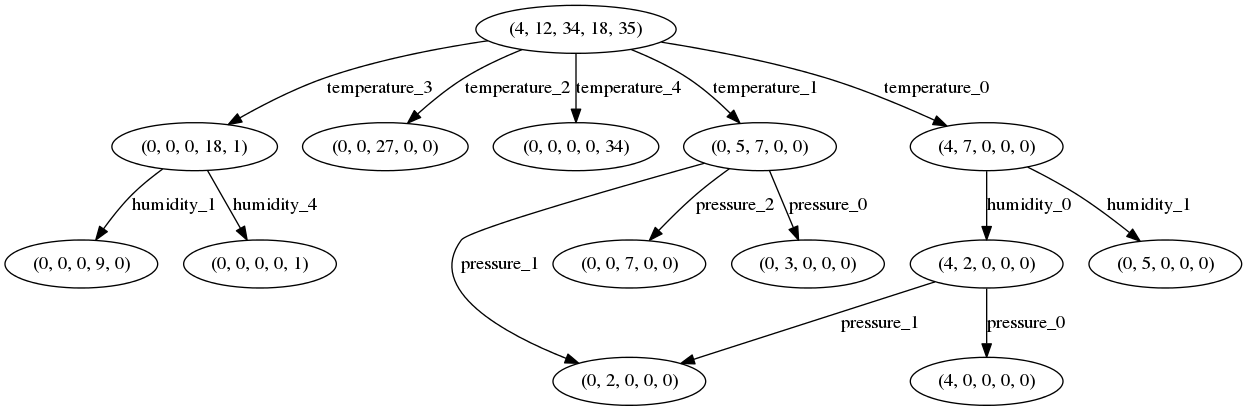}
\caption{Example of a tree in the random forest}
\label{tree}
\end{figure}

The random forest, constituted in this set of experiments by 100 trees, is defined as follows.
For each tree $T_i, i=1..100$:

\begin{itemize}
\item A sample of 67\% of dates $\tau_1, \hdots, \tau_{67} \in \{1,...,100\}$ is extracted
\item The root of the tree $T_i$ is the tuple $\left(\sharp\{j \mid F^{\tau_j}=n, j=1..67\}\right)_{n=1..5}$, where $\sharp X$ is the cardinality of the finite set $X$. Thus, its $n$-th coordinate corresponds to the number of times the device has been in the global failure $n$ in this sample of observation dates.
\item The category $c$ having the largest Gain for the dates in the root node is selected. The dates are divided into five sets depending on thresholds related to $c$. Then, 5 edges labeled by both $c$ and failure levels $l_i^0 = \{1,..,5\}$ are added to $T_i$, as depicted in Figure~\ref{tree}. They are directed to (at most) 5 new vertices containing the tuples $$\left(\sharp \{j \mid F^{\tau_j}=n \text{ and } d_i^{\tau_j^i}  \text{ has a } c \text{ level equal to } l_i \} \right)_{n=1,..,5}.$$
In other words, we only consider in this node a sub-sample of dates having their functioning level for category $c$ equal to $l_i^0$, and we divide the sub-sample into 5 subsets, depending on their global functioning levels: the tuple is constituted by each cardinality of these subsets, see Fig.~\ref{tree}.
\item The process is continued, with: this vertex as a new root, the reduced set of observed dates, and the categories minus $c$. It is stopped when either all the categories have been regarded, or when tuple of the node has at least 4 components equal to 0.
\end{itemize}

\subsection{Providing a diagnostic on a new set of observations}

Finally, given a new set of observations at a given time, the diagnostics of the industrial device is obtained as follows. 

Let $T$ be a tree in the forest. T will be visited starting from its root until reaching a leaf as described below.

\begin{enumerate}
\item All the edges connected to the root of $T$ are labeled with the same category $c$, but with various failure levels. The selected edge $e$ is the one whose labeled level of failure regarding $c$ corresponds to the $c$-level of failure of the observations.
\item If the obtained node $n$ following edge $e$ is a leaf, then the global level of failure of the observations according to $T$ is the coordinate of the unique non zero component of the tuple. If not, the tree walk is continued at item 1 with node $n$ as new root.
\end{enumerate}

The global diagnostics for the given observation is a majority consensus of all the responses of all the trees in the forest.

\subsection{Numerical simulations}

The training set is obtained by simulating $100$ observations for $10$ successive times, which results in $1000$ instances. The resulting data base is then used to train $100$ trees that will constitute the trained random forest. 

\begin{figure}[!ht]
\centering
\includegraphics[scale=0.275]{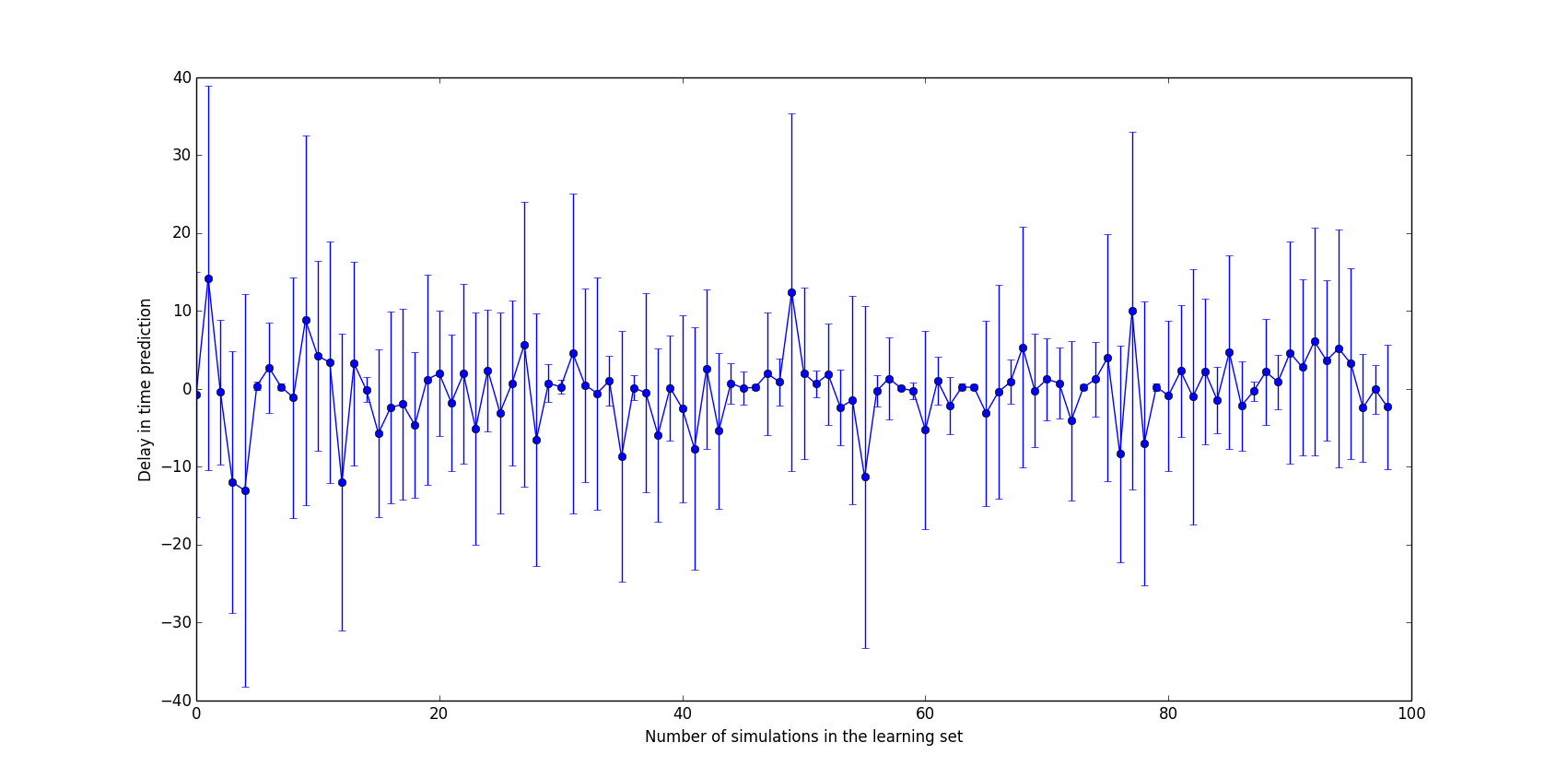}
\caption{Delay in failure detection with respect to the number of simulations.}
\label{delnocor}
\end{figure}

Figure \ref{delnocor} presents the delay between the time the system enters a failure mode and the time of its detection. This is done in the absence of correlations between the different features. The $0$ time value of delay, the negative values, and positive value refer to in time predictions, early predictions and late predictions of failures, respectively. The plotted values are the average result per number of simulations which varies from $1$ to $100$. With time, sensor nodes start to fail in order to simulate missing data packets. As a result, the RF algorithm was able to detect $54\%$ of the failures either in time or before their occurrence.

\begin{figure}[!ht]
\centering
\includegraphics[scale=0.275]{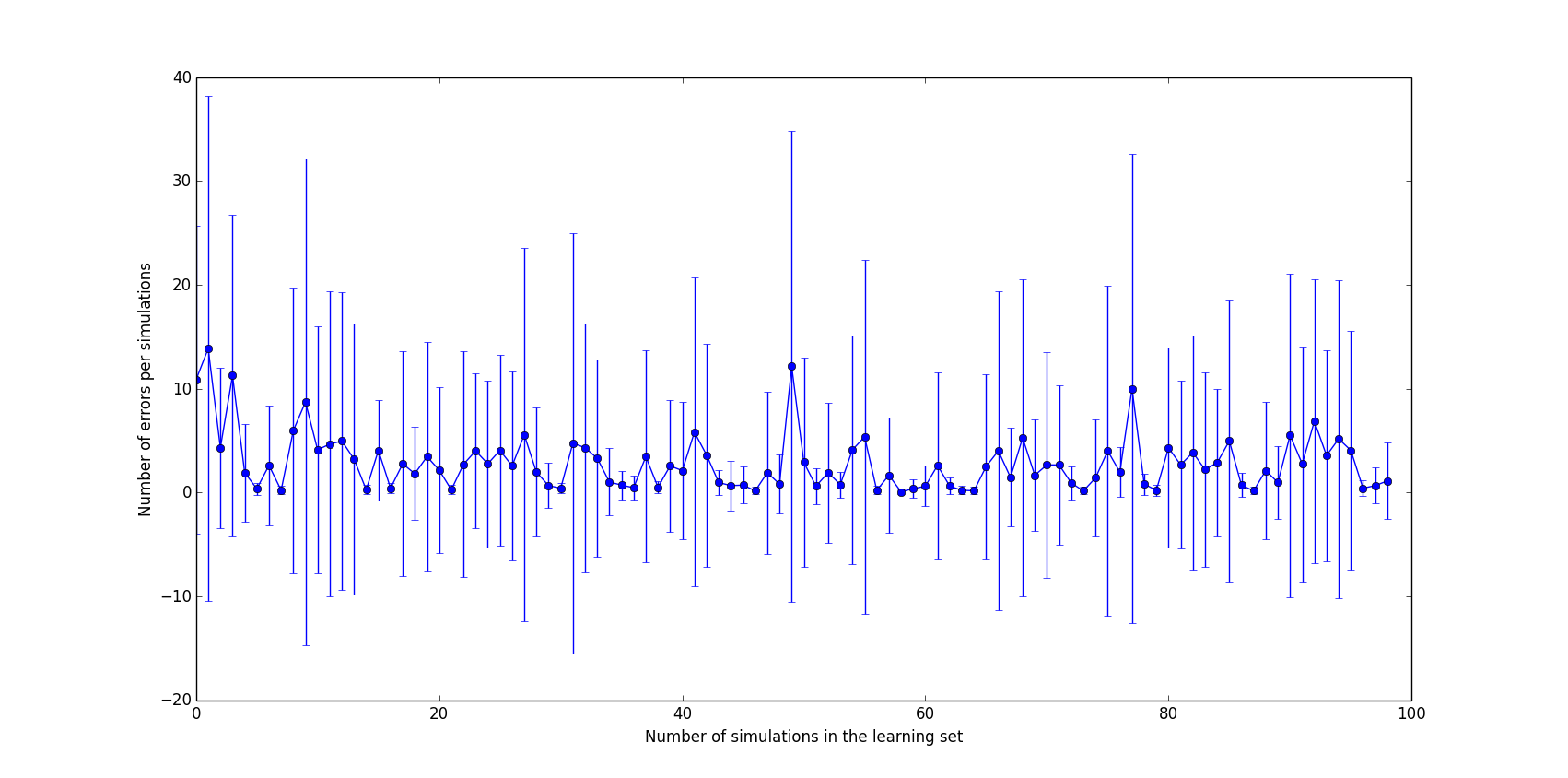}
\caption{Error rate in diagnostics with respect to the number of simulations}
\label{errnocor}
\end{figure}

For each of the $100$ performed simulations, we calculated the average number of errors in fault detection, produced by the trees in the forest. Figure \ref{errnocor} shows that this error rate remained below $15\%$ through the simulation. This error rate includes both \enquote{too early} and \enquote{too late} detections. When certain sensor nodes stop functioning, this leads to a lack on information, which has an impact on the quality of predictions; this explains a sudden increase in the error rate with time. We can conclude from the low error rate in the absence of some data packets that increasing the number of trees in the RF helps improve the quality and accuracy of predictions.

\begin{figure}[!ht]
\centering
\includegraphics[scale=0.5]{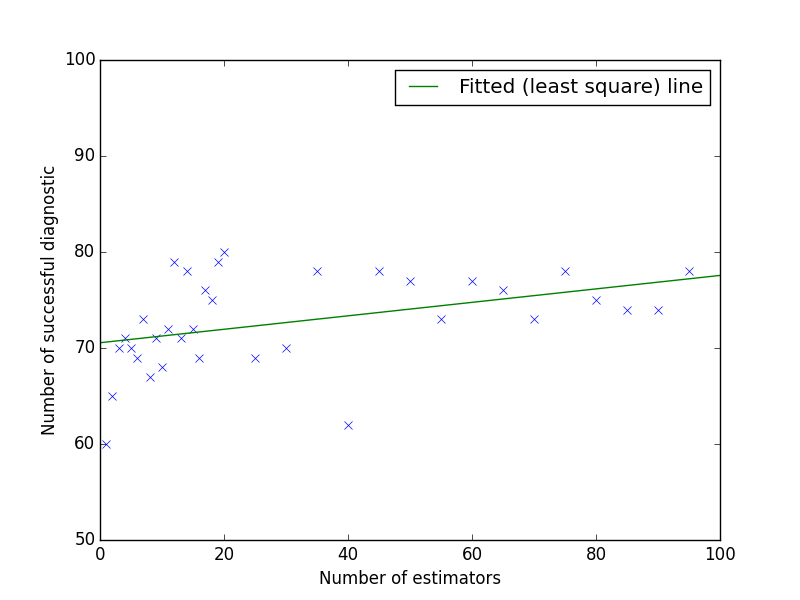}
\caption{Number of successful diagnostics with respect to the number of trees.}
\label{result}
\end{figure}

As described in Section \ref{set2}, a correlation was introduced between the features. Figure \ref{result} shows the number of successful diagnostics when the number of tree estimators in the forest changes. As shown in this figure, the RF method guarantees a $60\%$ success rate when the number of trees is limited to $5$. As this number grows, the accuracy of the method increases to reach $80\%$ when the number of trees is around $100$.

\section{Conclusion}
\label{conc}

Instead of using wired sensor networks for a diagnostics and health management method, it is possible to use wireless sensors. Such a use can be motivated by cost reasons or due to specific particularities of the monitored device.
In the context of a changing number and quality of provided features, the use of random forests may be of interest. These random classifiers were recalled with details in this article, and the reason behind their use in the context of a wireless sensors network monitoring was explained. 
Finally, algorithms and first examples of use of these random forests for diagnostics using a wireless sensor network were provided. The simulation results showed that the algorithm guarantees a certain level of accuracy even when some data packets are missing.

In future work, the authors' intention is to compare various tools for diagnostics to the random forests, either when considering wireless sensor networks or wired ones.
Comparisons will be carried out both theoretical and practical aspects.
The algorithm of random forests, for its part, will be extended to achieve prognostics and health management too. Finally, the method for diagnosing an industrial device will be tested on a life size model, to illustrate the effectiveness of the proposed approach.

\bibliographystyle{plain}
\bibliography{biblio}

\begin{thebibliography}{10}

\bibitem{Amit97}
Yali Amit and Donald Geman.
\newblock Shape quantization and recognition with randomized trees.
\newblock {\em Neural Computation}, 9:1545--1588, 1997.

\bibitem{Breiman96}
Leo Breiman.
\newblock Bagging predictors.
\newblock {\em Machine Learning}, 24:123--140, 1996.

\bibitem{Breiman99}
Leo Breiman.
\newblock Using adaptive bagging to debias regressions.
\newblock Technical report, Statics Department UCB, 1999.

\bibitem{Breiman01}
Leo Breiman.
\newblock Random forests.
\newblock {\em Machine Learning}, 45:5--32, 2001.

\bibitem{Dash00}
Sourabh Dash and Venkat Venkatasubramanian.
\newblock Challenges in the industrial applications of fault diagnostic
  systems.
\newblock {\em Computers and Chemical Engineering}, 24:785--791, 2000.

\bibitem{Dietterich00}
Thomas~G. Dietterich.
\newblock An experimental comparison of three methods for constructing
  ensembles of decision trees: Bagging, boosting, and randomization.
\newblock {\em Machine Learning}, 40:139--157, 2000.

\bibitem{Freund96}
Y.~Freund and R.~Schapire.
\newblock Experiments with a new boosting algorithm.
\newblock In {\em Proceedings of the Thirteenth International Conference on
  Machine Learning}, pages 148--156, 1996.

\bibitem{Fumera05}
Giorgio Fumera and Fabio Roli.
\newblock A theoretical and experimental analysis of linear combiners for
  multiple classifier systems.
\newblock {\em IEEE Transactions on Pattern Analysis and Machine Intelligence},
  27(6):942--956, 2005.

\bibitem{Ho98}
Tin~Kam Ho.
\newblock The random subspace method for constructing decision forests.
\newblock {\em IEEE Transactions on Pattern Analysis and Machine Intelligence},
  20(8):832--844, 1998.

\bibitem{Kanemoto13}
Shigeru Kanemoto, Norihiro Yokotsuka, Noritaka Yusa, and Masahiko Kawabata.
\newblock Diversity and integration of rotating machine health monitoring
  methods.
\newblock In {\em Chemical Engineering Transactions}, number~33, pages
  169--174, Milan, Italy, 2013.

\bibitem{Moghaddass14}
Ramin Moghaddass and Ming Zuo.
\newblock An integrated framework for online diagnostic and prognostic health
  monitoring using a multistate deterioration process.
\newblock {\em Reliability Engieneering and System Safety}, 124:92--104, 2014.

\bibitem{Schapire99}
Robert~E. Schapire.
\newblock A brief introduction to boosting.
\newblock In {\em Proceedings of the sixteenth International Joint Conference
  on Artificial Intelligence}, 1999.

\bibitem{Sharkey97}
A.~Sharkey and N.~Sharkey.
\newblock Combining diverse neural nets.
\newblock {\em The Knowledge Egineering Review}, 12(3):231--247, 1997.

\bibitem{Tsymbal00}
Alexey Tsymbal and Seppo Puuronen.
\newblock Bagging and boosting with dynamic integration of classifiers.
\newblock In {\em The 4th European Conference on Principles and Practice of
  Knowledge Discovery in Data Bases PKDD}, pages 116--125, 2000.

\bibitem{Tumer96}
K.~Tumer and J.~Ghosh.
\newblock Error correlation and error reduction in ensemble classifiers.
\newblock {\em Connection Science}, 8:385--404, 1996.

\bibitem{Verikas11}
A.~Verikas, A.~Gelzinis, and M.~Bacauskiene.
\newblock Mining data with random forests: A survey and results of new tests.
\newblock {\em Pattern Recognition}, 44:330--349, 2011.

\end{thebibliography}

\end{document}